\documentclass{fdibaconf}

\usepackage{lipsum}
\usepackage{cite}
\usepackage{amsmath,amssymb,amsfonts}
\usepackage{algorithmic}
\usepackage{graphicx}
\usepackage{textcomp}

\usepackage{framed}   		
\usepackage{xcolor}   		
\usepackage{chngcntr} 		
\usepackage{bookmark} 		
\usepackage{mdframed} 		

\usepackage{caption}
\usepackage{subcaption}

\usepackage{pgfplots}
\pgfplotsset{compat=newest}
\usepackage{tikz}
\usepackage{tikzscale}
\usepackage{placeins}
\usepackage{dblfloatfix}
\usetikzlibrary{external}
\newcommand{\eg}{e.\,g.~}

\newcommand{\wrt}{w.\,r.\,t.~}

\newcommand{\vek}[1]{{\mathbf #1}}

\newcommand{\sv}[1]{\boldsymbol{#1}}

\newcommand{\play}[1]{^{(#1)}}

\hypersetup{%
	urlcolor=black,
	citecolor=black,
	linkcolor=black,
	hidelinks=false
}

\usepackage{multicol}
\usepackage{fancyhdr}
\fancypagestyle{firstpage}{
	\fancyhf{}
	\fancyhead[C]{\vspace*{3mm}This work has been submitted and accepted for a publication of 2022 FDIBA Conference "Engineering 4.0 and The Internet of Everything" in Sofia, Bulgaria. Copyright may be transferred without notice, after which this version may no longer be accessible.}
}
\begin{document}
\pagenumbering{gobble} 
\pagestyle{empty} 
\title{Entwicklung eines mobilen Arbeitsmaschine-Simulators für die Validierung kooperativer Regelungskonzepte} 

\titel{Development of a Mobile Vehicle Manipulator Simulator for the Validation of Shared Control Concepts} 

\author{
\authorname{Balint Varga\addressmark[1], Selina Meier\addressmark[1] and S\"oren Hohmann\addressmark[1]}\\
\address{
\addressmark[1]Institute of Control Systems (IRS) at the Karlsruhe Institute of Technology (KIT),\\
76131 Karlsruhe, Germany, e-mail: balint.varga2@kit.edu\\
}}

\abstracttext{
This paper presents the development of a real-time simulator for the validation of controlling a large vehicle manipulator. The need for this development can be justified by the lack of such a simulator: There are neither open source projects nor commercial products, which would be suitable for testing cooperative control concepts. First, we present the nonlinear simulation model of the vehicle and the manipulator. For the modeling MATLAB/Simulink is used, which also enables a code generation into standalone C++ ROS-Nodes (\textit{Robot Operating System} Nodes). The emerging challenges of the code generation are also discussed. Then, the obtained standalone C++ ROS-Nodes integrated in the simulator framework which includes a graphical user interface, a steering wheel and a joystick. This simulator can provide the real-time calculation of the overall system's motion enabling the interaction of human and automation. Furthermore, a qualitative validation of the model is given. Finally, the functionalities of the simulator is demonstrated in tests with a human operators.
}
{
In diesem Beitrag wird die Entwicklung eines echtzeitfähigen Simulators für die Validierung der Regelung einer großen mobilen Arbeitsmaschine vorgestellt. 
Die Notwendigkeit dieser Entwicklung lässt sich mit dem Fehlen eines solchen Simulators begründen: Es gibt weder Open-Source-Projekte noch kommerzielle Produkte für solche Systeme, die zum Testen von kooperativen Regelungskonzepte geeignet wären. Zunächst wird das nichtlineare Simulationsmodell des Fahrzeugs und des Manipulators vorgestellt. Für die Modellierung wird MATLAB/Simulink verwendet, das auch eine Codegenerierung in standalone C++ ROS-Nodes (\textit{Robot Operating System} Nodes) ermöglicht. Die resultierenden Herausforderungen der Codegenerierung werden ebenfalls diskutiert. Anschließend werden die standalonen C++ ROS-Nodes in ein Framework integriert, das aus einer grafischen Benutzeroberfläche, einem Lenkrad und einem Joystick besteht. 
Der Simulator ist in der Lage, die Bewegung des Gesamtsystems in Echtzeit zu berechnen, wodurch die Interaktion zwischen Mensch und Automation ermöglicht wird. Außerdem wird eine qualitative Validierung des Modells vorgenommen. Schließlich wird die Funktionsweise des Simulators in Tests mit einem menschlichen Bediener demonstriert.
}

\maketitle

\selectlanguage{german}
\thispagestyle{firstpage}
 
\section{Introduction}
\begin{figure}[t!]
	\centering
	\includegraphics[width=0.99\linewidth]{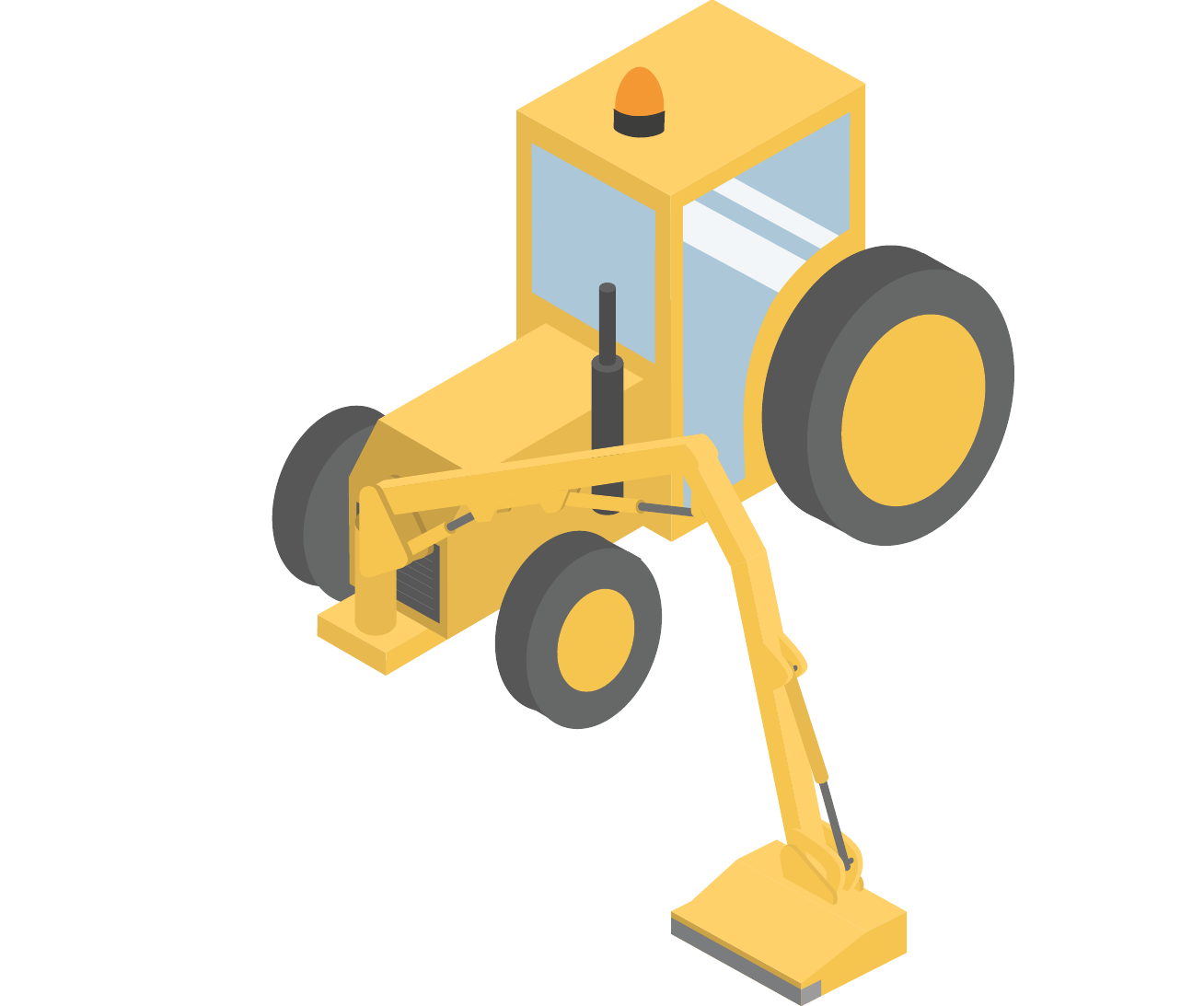}
	\caption{Die Skizze des Simulationsmodells, welches ein Fahrzeug (Traktor, Lastkraftwagen o.ä.) und einen hydraulischen Manipulator mit vier Gelenken enthält.}
	\label{fig:sim_model}
\end{figure}

Große Arbeitsmaschinen (siehe Abb. \ref{fig:sim_model}) werden in verschiedenen Bereichen eingesetzt, z. B. in der Landwirtschaft \cite {2020_ExtensiveReviewMobile_fue}, der Forstwirtschaft \cite{2021_AdvancesForestRobotics_oliveira} oder der Straßeninstandhaltung \cite {2017_EffectsChangeOrders_shrestha}. Solche Systeme werden in der Regel von einem menschlichen Bediener gesteuert, da die Arbeitsumgebung unstrukturiert ist. Abb. \ref{fig:sim_model} zeigt eine große Arbeitsmaschine mit einem am Fahrzeug angebrachten Manipulator.

Normalerweise muss der Bediener mit dem Manipulator verschiedene Aufgaben ausführen und gleichzeitig das Fahrzeug auf der Straße halten. Diese zwei Aufgaben sind für den Menschen geistig und manchmal auch körperlich fordernd. Die vollständige Automatisierung einer großen Arbeitsmaschine ist aufgrund von Sicherheitsvorschriften und der komplexen Arbeitsumgebung des Manipulators nicht möglich, siehe \cite{2017_EffectsChangeOrders_shrestha}, \cite{2018_MotionControlMultiactuator_xu}.

Um den Bediener zu entlasten, wurden kooperative Regelungskonzepte entwickelt: Bei diesen Arbeiten wurde die Fahrzeugführung so automatisiert, dass der Bediener mit dem Manipulator optimal arbeiten kann und er sich nicht auf das Fahrzeug konzentrieren muss, siehe \cite{2019_ControlLargeVehicleManipulators_varga}, \cite{2020_LimitedInformationCooperativeShared_varga} and \cite{2022_LimitedInformationShared_varga}. 
Allerdings gibt es keinen Simulator, welche sich für das Testen solcher kooperativen Regelungskonzepte eignen würde. Deswegen ist das Ziel dieser Arbeit, einen Simulator zu entwickeln, der sich für solche Validierungsvorhaben eignet.

Dieser Beitrag ist wie folgt strukturiert: In dem Abschnitt~II. werden die entwickelten Teilsysteme präsentiert. Abschnitt III. befasst sich mit dem qualitativen Validierungsszenario. Danach werden in dem Abschnitt IV. die Ergebnisse diskutiert. Anschließend fasst Abschnitt V. den Beitrag zusammen.

\section{Entwicklung der Teilsysteme} \label{sec:teilsystems}
Ein \glqq out-of-shell \grqq{} Gesamtsimulationsmodell, das sich für die Verifizierung und Validierung der vorgeschlagenen gemeinsamen Steuerungskonzepte eignen würde, existiert in der Literatur nicht. Dennoch gibt es im Stand der Technik Modelle der Teilsysteme, die zum Aufbau der Gesamtsimulation verwendet werden können: Es gibt Modelle von Schwerlastfahrzeugen und großen hydraulischen Manipulatoren. Ihre Kombination ist jedoch neuartig und wirft weitere Herausforderungen auf. Außerdem erfordert die Echtzeitimplementierung\footnote{Eine Echtzeitfähigkeit ist notwendig, um kooperative Regelungskonzepte mit einem Menschen in Experimenten untersuchen zu können.} zusätzliche Untersuchungen. Das hier vorgestellte Modell wurde in \cite{2019_ModelPredictiveControl_varga} teilweise verwendet, aber eine detalliertere Untersuchung und eine Weiterentwicklung sind notwendig.

\subsection{Das Modell des Fahrzeugs}
Die Abb. \ref{fig:sim_model} zeigt das Gesamtsystem: Ein mittelgroßes Schwerlastfahrzeug und ein großer hydraulischer Manipulator.
Das Simulationsmodell des Fahrzeugs 
\begin{figure}[h!]
	\normalsize
	\centering
	\includegraphics[width=0.99\linewidth]{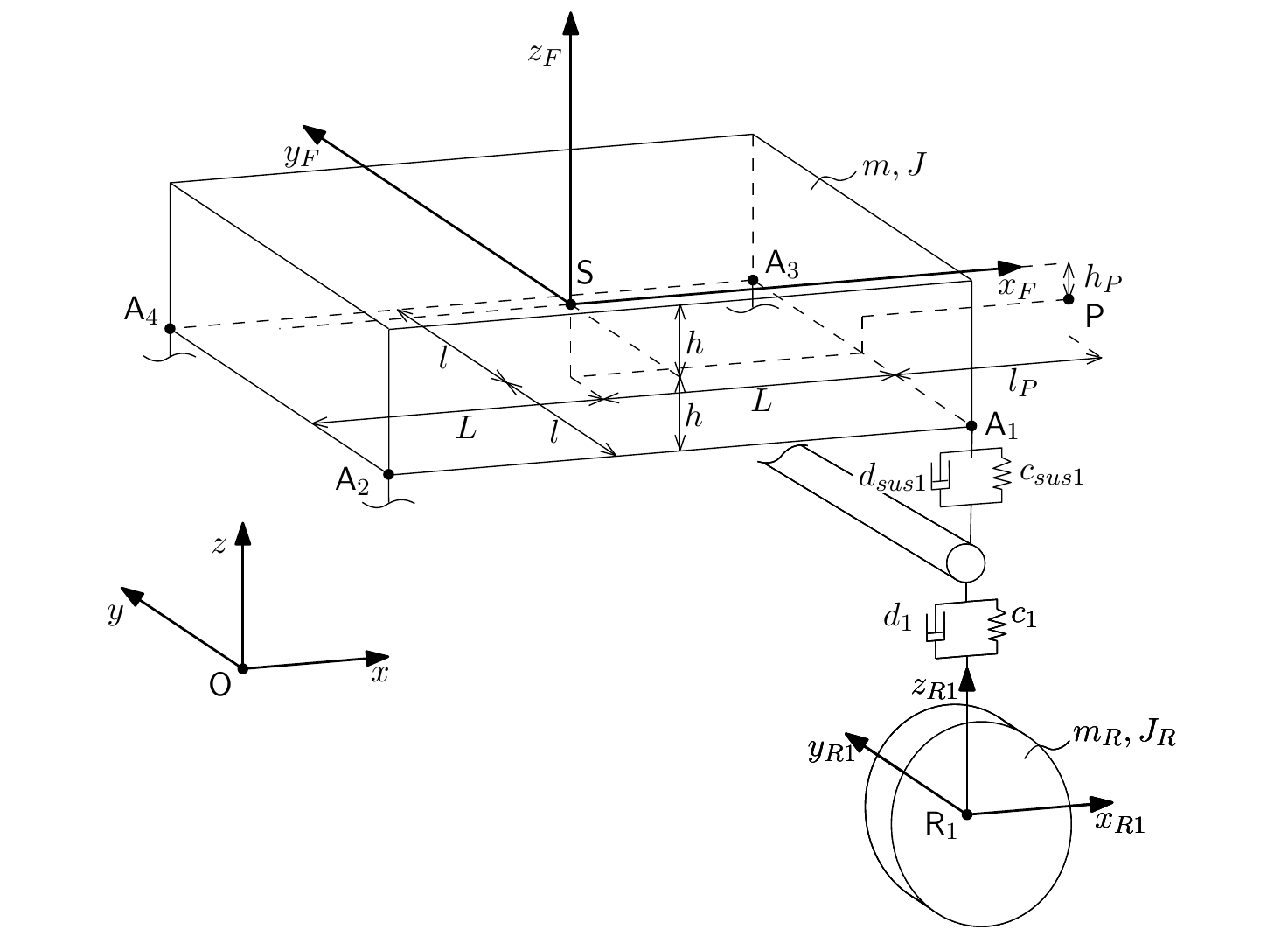}
	\caption{Das nichtlineare Modell des Fahrzeugs}		
	\label{fig:veh_dyn_model}
\end{figure}
enthält ein dreidimensionales Fahrgestell, zwei zweidimensionale Aufhängungen und vier Räder mit einem linearen Reifenmodell (Abb.~\ref{fig:veh_dyn_model}). 
Dieses Modell kann komplexere Bewegungen beschreiben, die nicht mit denen von Personenkraftwagen vergleichbar sind (z.B. größere Roll- und Nickwinkel). Da die Bewegung des Fahrzeugs keine hohe Dynamik aufweist, ist ein lineares Reifenmodell ausreichend. Um die Echtzeitfähigkeit des Simulators zu gewährleisten, wird das Fahrwerksbewegungsmodell aus \cite{2014_ModelingCommercialVehicles_kovacs} implementiert.
\begin{figure}[b!]
	\centering
	\includegraphics[width=0.99\linewidth]{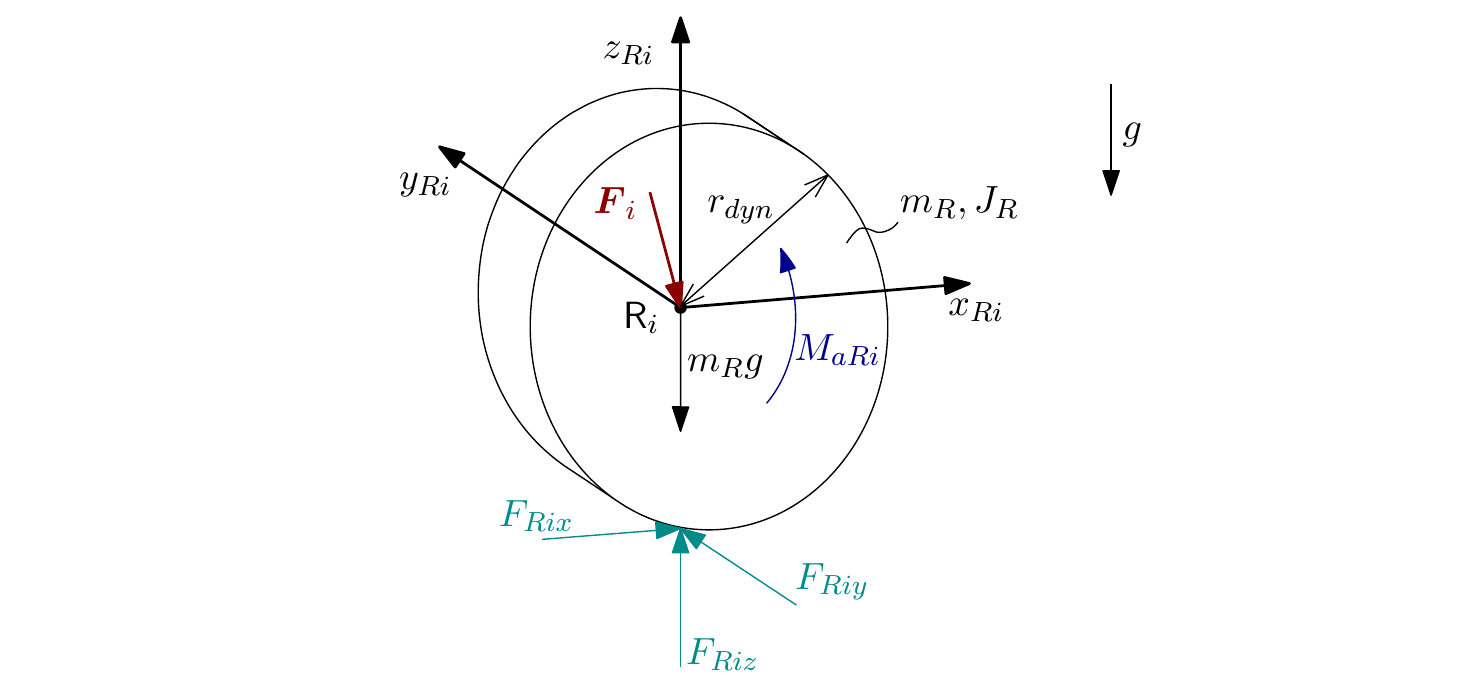}
	\caption{Der Freischnitt eines Rades für die Herleitung der Bewegungsgleichungen.}
	\label{fig:rad_model}
\end{figure}
Für die Herleitung der Radbewegungen werden die Kräfte mithilfe der Abb. \ref{fig:rad_model} aufgestellt. Die Winkelbeschleinigung eines Rades wird wie folgt berechnet
\begin{equation}
\dot{\omega}_\mathrm{Rad} = \frac{1}{\theta_\mathrm{Rad}} \left( F_\mathrm{Rix} \cdot r_\mathrm{Rad} + M_\mathrm{aRi} \right).
\end{equation}
Dabei muss die Eigendynamik der Räder um Geschwindigkeit $0\,$m/s berücksichtigt werden, weil die Kraft 
$
F_\mathrm{Rix} = c_i \cdot s_i
$ 
aus dem Schlupf ($s_i$) berechnet wird. Der Parameter $c_i$ ist die Steifigkeit des linearen Reifenmodels, $M_\mathrm{aRi}$ ist der Antriebsmoment des Rades. Die Herausforderung besteht darin, dass die Berechnung des Schlupfs mit diskreten Solvern um $v=0\,$m/s numerisch instabil sein kann. Dies würde die weitere Nutzung des Simulationsmodells behindern. Die vorgeschlagene Lösung ist die Einführung eines numerischen Schwellenwerts für die Schlupfberechnung, siehe \cite[Kapitel 4.]{rill2011road}. 
Daher wird der Schlupf des Rades berechnet als 
\begin{equation}
	s = \frac{v_\mathrm{Fahr} - \omega_\mathrm{Rad} \cdot r_\mathrm{Rad}}{\mathrm{max}\left(v_\mathrm{Fahr}, \omega_\mathrm{Rad}\cdot r_\mathrm{Rad}, v_\mathrm{N}\right)},
\end{equation}
wobei $v_\mathrm{Fahr}$, $\omega_\mathrm{Rad}$ und $r_\mathrm{Rad}$ die Fahrzeuggeschwindigkeit, die Drehgeschwindigkeit des Rades bzw. der Radius des Rades sind. Der Parameter $v_\mathrm{N}$ wird so gewählt, dass die numerische Stabilität bei niedrigen Geschwindigkeiten erhalten bleibt. 
Die mobile Arbeitsmaschine wird in einem globalen Koordinatensystem modelliert, die berechnete Position der Subsysteme wird sowohl im globalen Rahmen als auch relativ zum Fahrzeugchassis angegeben. 

\subsection{Das Modell des hydraulischen Manipulators}
Der hydraulische Manipulator besteht aus drei Armsegmenten und vier Gelenken, siehe Abb. \ref{fig:sim_model} und sein Zustandsvektor ist $\sv{\varphi} = \left[\varphi_1, \; \varphi_2, \; \varphi_3, \; \varphi_4 \right]$. Die Modellierung der schweren Armsegmente und der hydraulischen Aktoren basiert auf \cite[Kapitel 6.]{2015_Oelhydraulik_findeisen} \cite{2017_FullReducedorderModel_ruderman} und \cite{2017_IdentificationControlDesign_rudolfsen}. Zusätzlich wird eine geschwindigkeitsbasierte inverse Kinematik hinzugefügt. Dies bedeutet, dass der Bediener nicht direkt die Gelenke des Manipulators mit dem Joystick steuert, sondern die Geschwindigkeiten des Endeffektors. Die allgemeinen Bewegungsgleichungen lauten
\begin{align} \label{eq:manip_model_simpl} \nonumber
	\vek{M} \ddot{\sv{\varphi}}(t) &= \sv{T}_\mathrm{hyd} \left(\vek{p}_\mathrm{hyd}, \sv{x}_\mathrm{hyd}(t), \sv{u}\play{\mathrm{h}}(t,\sv{\varphi}) \right)  \\ \nonumber
	&+ \sv{T}_\mathrm{fric} \left(\dot{\sv{\varphi}}(t), \sv{\varphi}(t) \right)\\
	& + \sv{T}_\mathrm{mech}\left(\vek{p}_\mathrm{geo}, \sv{x}_\mathrm{veh,ext}(t) \right),
\end{align}
wobei $\vek{M}$ die Massenmatrix ist. Auf der rechten Seite von (\ref{eq:manip_model_simpl}) sind die Antriebsmomente des hydraulischen Aktuators ($\sv{T}_\mathrm{hyd}$), die linearen und nichtlinearen Reibungsmomente ($\sv{T}_\mathrm{fric}$) bzw. die externen mechanischen Einflüsse ($\sv{T}_\mathrm{mech}$) zu finden. Die Antriebsmomente des hydraulischen Aktuators hängen vom hydraulischen Parameter $\vek{p}_\mathrm{hyd}$ und von der Eingabe des Menschen $\sv{u}\play{\mathrm{h}} (\sv{\varphi})$ ab. Die internen Zustände des Hydraulikaktuators $\sv{x}_\mathrm{hyd}(t)$ sind der Öldruck, der Öldurchfluss und der Zustand der Elektromotoren, welche ie die Ventilatoren für die Hydraulikzylinder antreiben. Die modellierten Reibungen umfassen eine lineare viskose Komponente und eine Stribeck-Kurve unter Verwendung des LuGre-Reibungsmodells, welche ein gängiger Ansatz für Hydrauliksysteme ist, siehe \eg \cite{2012_ModelingDynamicFriction_tran}. Der weitere Einfluss der Bewegung des Fahrzeugs und der durch die Konfiguration des Manipulators verursachten Drehmomente/Kräfte. In (\ref{eq:manip_model_simpl}) steht $\sv{x}_\mathrm{veh,ext}$ für den erweiterten Fahrzeugzustand und $\vek{p}_\mathrm{geo}$ enthält die geometrischen Parameter, die für die Bewegungserzeugung des Manipulators relevant sind. Die Parameter des hydraulischen Manipulators wurden aus der Literatur entnommen, \cite{2009_ExperimentalEvaluationBilateral_zarei-nia} und \cite{2012_ModelingSimulationHighPerformance_alaydi}. 

Die beiden Teilsysteme sind in einem Simulink-Modell implementiert. Das Gesamtsimulationsmodell läuft mit $2\,$kHz Abtastrate und mit einem Runge-Kutta-Solver. Der Solver wird gewählt, um die numerische Stabilität des Gesamtsystems zu gewährleisten: Das Ölmodell der Hydraulikzylinder weist bei Solvern niedrigerer Ordnung hochfrequente numerische Schwingungen auf, was der Grund für die Wahl des Solvers war.
Mit diesen Einstellungen ist das Gesamtmodell in der Lage, numerisch stabile Bewegungen zu generieren.

\begin{figure*}[t!]
	\normalsize
	\centering
	\includegraphics[width=0.95\linewidth]{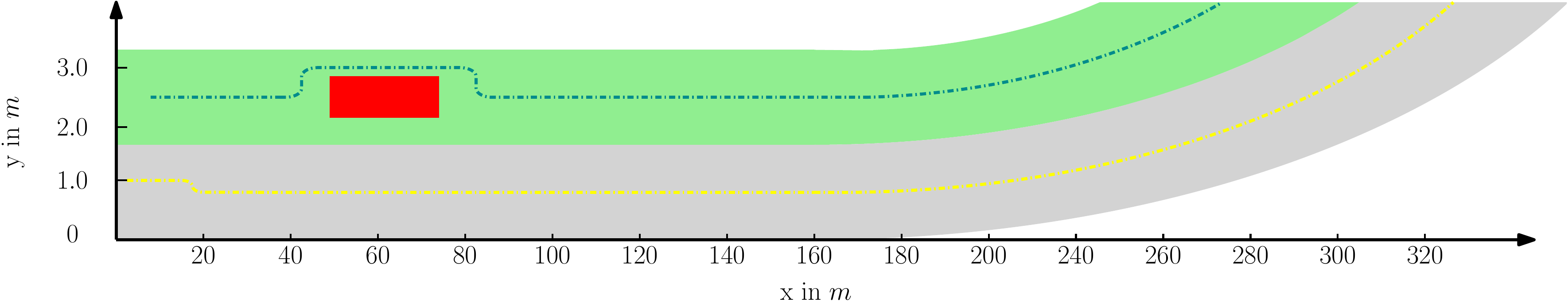}
	\caption{Die schematische Darstellung des Validierung Szenario mit den Referenztrajektorien der mobilen Arbeitsmaschine. Es ist zu beachten, dass die Skalierung der Achsen $x$ und $y$ aufgrund der besseren Visualisierung des Szenarios unterschiedlich ist.}		
	\label{fig:vm_scenario_ref}
\end{figure*}
\section{Das Szenario der qualitativen Validierung}
Der Mensch in der Simulation ist als Optimalregler modelliert. Dies bedeutet, dass er seine Eingaben mit der Optimierung einer Gütefunktion berechnet. Die Stellgrößen des Menschen werden über

\begin{subequations}
	\begin{align} \label{eq:hum_cost_fisc}
		{\sv{u}\play{h}}^*(t) &= \underset{\sv{u}\play{h}}{\mathrm{arg \, min}}\, J\play{h}\left(\sv{x}(t),\tau_\mathrm{end},\sv{u}\play{h}(t)\right) \\
		\mathrm{\wrt}& \; \sv{f}\left(t,\sv{x}(t),\sv{u}\play{h}(t) \sv{u}\play{\mathrm{a}}(t)\right)\; \; t \in [0,\tau_\mathrm{end}] \\ \nonumber
		&  \sv{x}(0) = \sv{x}_0,
	\end{align}
\end{subequations}
bestimmt. Diese dynamische Optimierung setzt sich aus der Kostenfunktion des Menschen\footnote{Diese Kostenfunktion ermöglicht die Modellierung der Ziele und Präferenzen des Menschen. Dadurch ist eine vielseitige Beschreibung und eine anwendungsspezifische Anpassungen des Modells realisierbar.} und der Dynamik des technischen Systems, $\sv{f}(\cdot)$ zusammen. Die Dynamik wird durch die Eingabe des Menschen ($\sv{u}\play{\mathrm{h}}(t)$) und der Automatisierung ($\sv{u}\play{\mathrm{a}}(t)$) beeinflusst, wodurch komplexere Bewegungen entstehen können. In der Literatur wurde gezeigt, dass ein lineares Systemmodell und eine quadratische Kostenfunktion (LQ-Probleme) viele Anwendungen gut genug beschreiben können. Deswegen ist die menschliche Kostenfunktion
\begin{align*} 
		J\play{h}&=\frac{1}{2}\int_{0}^{\tau_\mathrm{end}}  \sv{x}(t)^T  \vek{Q}\play{h}\sv{x}(t)  \\
		&+ {\sv{u}\play{h}(t)}^T \vek{R}\play{hh}\sv{u}\play{h}(t)  + {\sv{u}\play{a}(t)}^T \vek{R}\play{ha}\sv{u}\play{a}(t) \text{ d}t,
\end{align*}
wobei die Matrizen $\vek{Q}\play{h}\geq\vek{0}$, $\vek{R}\play{hh}>\vek{0}$ and $\vek{R}\play{ha}>\vek{0}$ die Bestrafungsfaktoren sind. Die für die Optimierung verwendete Systemdynamik wurde in \cite{2019_ModelPredictiveControl_varga} hergeleitet. 
Abb. \ref{fig:vm_scenario_ref} zeigt die schematische Darstellung des Validierung Szenario und die beiden Referenztrajektorien der mobilen Arbeitsmaschine. Zunächst hat das Fahrzeug eine kleine Korrektur in seiner Referenz bei $x=20\,$m. Sie kann durch kleine Änderungen im Verkehrsfluss verursacht werden, die eine Anpassung \eg für mehr Sicherheitsabstand zu anderen Fahrzeugen erfordern.
Um $x=45\,$m hat die Referenz des Manipulators einen plötzlichen Sprung. Dies modelliert ein verstecktes Hindernis (größerer Stein, Metallteil, siehe roter Block in der Abb.), das den Manipulator beschädigen könnte, worauf der Fahrer reagieren muss. 
Danach fährt die mobile Arbeitsmaschine in eine sanfte Kurve. Der plötzliche Schritt des Manipulators wirkt sich auf die Bewegung des Fahrzeugs aus, was sich an den Drehwinkeln ablesen lässt.

In dem folgenden Szenario wird das menschliche Modell für die Berechnung der Eingangssignale benutzt. Allerdings ist der Simulator so ausgelegt, dass ein Austausch des simulierten Menschen die anderen Komponenten nicht beeinflusst. Dadurch ist eine umfassende Untersuchung der verschiedenen Regelungskonzepte möglich.

\section{Ergebnisse und Diskussion} \label{sec:erg_disk}

\subsection{Fahrzeugbewegungen}
\begin{figure}[t!]
	\normalsize
	\centering
	\includegraphics[width=0.99\linewidth]{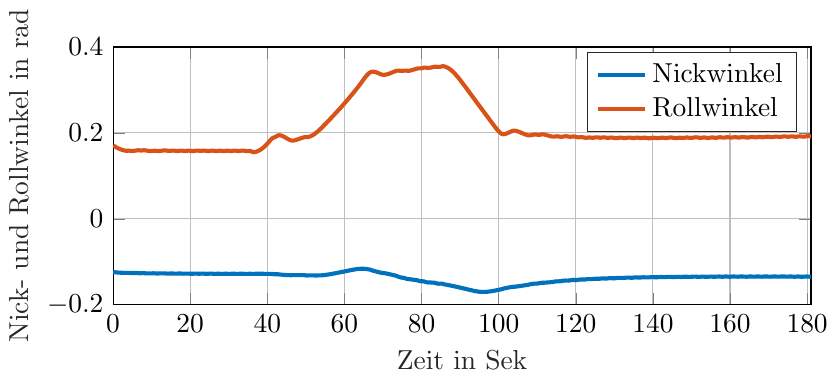}
	\caption{Die Nick- und Rollwinkel des Fahrzeugs. Der Einfluss der der Bewegung des Manipulators ist besonders um $t = 50\,$s zu sehen.}		
	\label{fig:veh_pitch_roll}
\end{figure}

Die Bewegung des Fahrzeugs weist keine gängigen Eigenschaften für Lastwagen oder Traktoren auf: Das Fahrzeug ist aufgrund der Masse des Manipulators um seine Längsachse gekippt. Abb. \ref{fig:veh_pitch_roll} zeigt, dass der Rollwinkel des Fahrzeugs größer als $0,087\,$rad (ca. $5\,$deg) ist. Daher ist ein lineares Fahrzeugmodell nicht ausreichend und die Implementierung dieses nichtlinearen Modells ist zweckmäßig, siehe~z.B.~\cite{2016_Analysis_forster}. Außerdem zeigt Abb.~\ref{fig:veh_pitch_roll} den dynamischen Einfluss des Manipulators: Um $x=60\,$s wird der Manipulator gestreckt, wodurch das Drehmoment auf das Fahrzeug größer wird, was zu einer Änderung des Rollwinkels führt. Nach $x=100\,$s bewegt sich der Manipulator wieder in die frühere Position zurück und das Fahrzeug wird \glqq zurückgekippt\grqq{}. Eine ähnliche Veränderung des Nickwinkels ist auch zu beobachten: Das Fahrzeug wird während des Manövers zwischen $60$ und $100\,$s \glqq nach vorne gekippt\grqq{}. Dies bestätigt die Notwendigkeit des nichtlinearen Fahrzeugmodells: Der Rollwinkel schwankt zwischen $0.2$ und $0.4$ rad, was größer als $0.087$ rad ist und bedeutet dass der Linearisierungsfehler zu groß wäre.
\begin{figure}[t!]
	\normalsize
	\centering
	\includegraphics[width=0.95\linewidth]{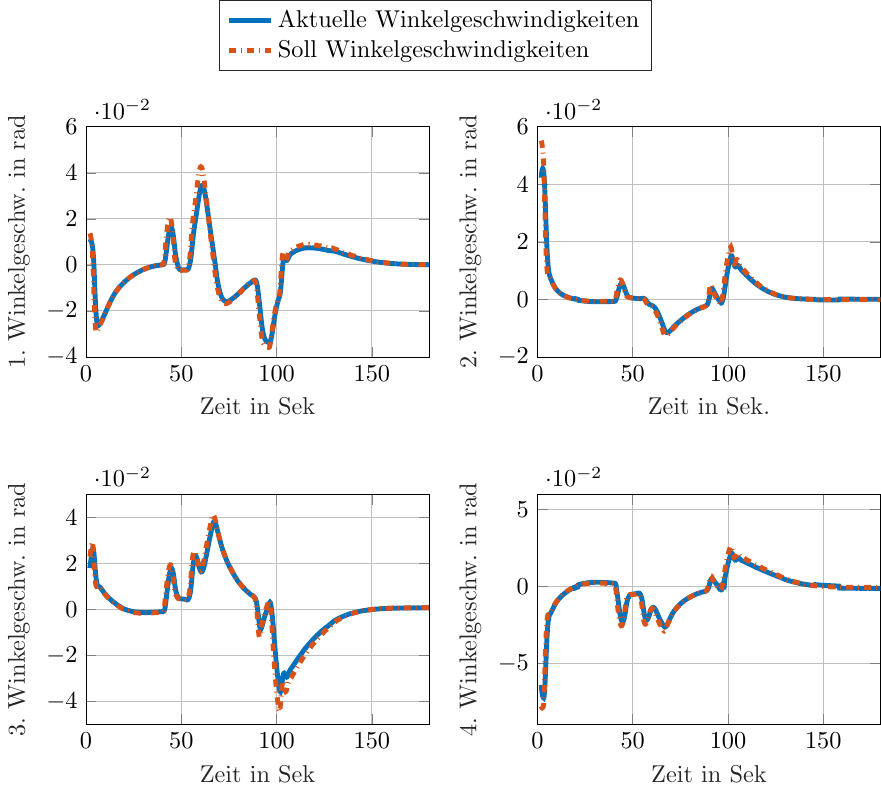}
	\caption{Die Soll- und Ist-Winkelgeschwindigkeiten der vier Gelenke. Ihre Bewegungen sind im Vergleich zu elektrischen Aktoren langsam und entsprechen den Ergebnissen aus der Literatur. Daher eignet sich das Modell des Manipulators für die Validierung.}		
	\label{fig:manip_joints_all}
\end{figure}

\begin{figure}[t!]
	\normalsize
	\centering
	\includegraphics[width=0.9\linewidth]{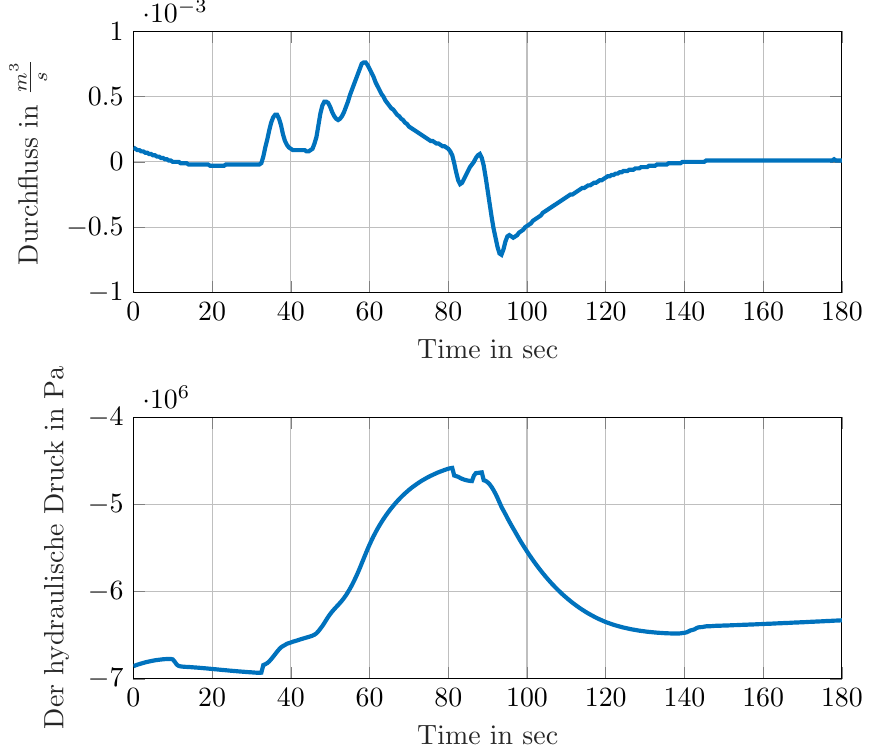}
	\caption{Der hydraulische Durchfluss und Druck des dritten Stellglieds des Manipulators.}		
	\label{fig:manip_hydarulics}
\end{figure}
\begin{figure*}[b!]
	\normalsize
	\centering
	\includegraphics{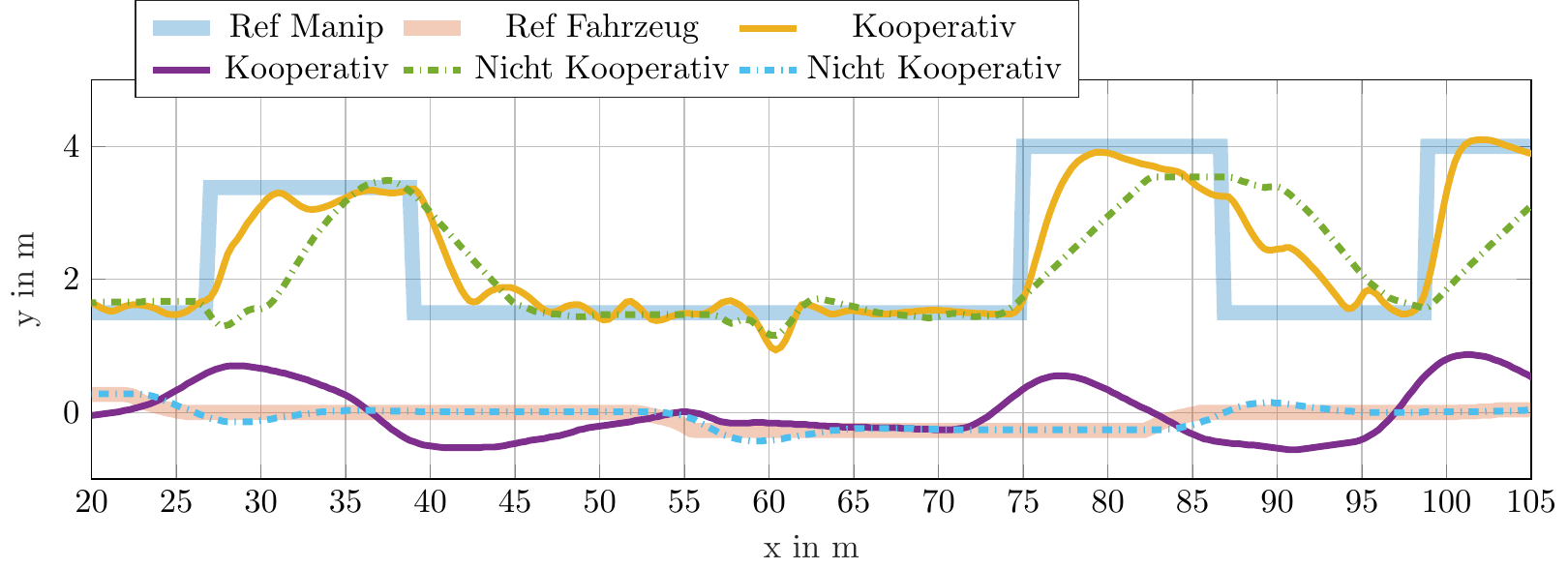}
	\caption{Der Vergleich der Trajektorien der kooperativen und der nicht kooperativen Regelung. Es ist ersichtlich, dass mit der entwickelten kooperativen Steuerung eine bessere Verfolgung der Trajektorien möglich ist.}		
	\label{fig:thms_validation}
\end{figure*}
\subsection{Bewegung des Manipulators}

Die Bewegungen von großen und hydraulisch angetriebenen Manipulatoren sind in der Regel wegen der großen Massenträgheit der Komponenten langsam und verzögert, vgl.~\cite{2017_IdentificationControlDesign_rudolfsen}. Die hydraulischen Aktuatoren können große Drehmomente abgeben, haben aber im Vergleich zu Elektromotoren eine langsame Dynamik. Der Bediener verwendet eine koordinierte Manipulatorsteuerung, was bedeutet, dass die gewünschten Geschwindigkeiten des Endeffektors, $\dot{x}_\mathrm{man}$ und $\dot{y}_\mathrm{man}$, koordiniert festgelegt werden. So kann der Bediener den Manipulator einfacher ansteuern.
Aus diesen Soll-Geschwindigkeiten werden die Soll-Winkelgeschwindigkeiten des Manipulators mit einem numerischen inversen kinematischen Algorithmus berechnet. Die gewünschten Winkelgeschwindigkeiten werden durch den Low-Level-Controller der hydraulischen Aktuatoren eingestellt. 

Für dieses Testszenario sind die Soll- und Ist-Winkelgeschwindigkeiten des Manipulators in Abb. \ref{fig:manip_joints_all} dargestellt. Es ist die langsame Bewegung des Manipulators zu erkennen. Diese Werte werden mit verfügbaren Ergebnissen aus der Literatur verglichen, was die Eignung und Genauigkeit des Modells zeigen. 
Um das modellierte Hydrauliksystem zu analysieren, zeigt Abb. \ref{fig:manip_hydarulics} den simulierten Ölfluss ($\dot{Q}_\mathrm{oil}$) und den Druck ($P_\mathrm{oil}$) im hydraulischen Aktuator des dritten Gelenks. Ihre Eigenschaften und Größen korrelieren auch mit den in der Literatur verfügbaren Ergebnissen. Es ist zu erkennen, dass während des plötzlichen Manövers der Druck und der Öldurchfluss erhöht werden, um das notwendige Drehmoment und die Kraft zur Bewegung des Gelenks bereitzustellen. In der sanften Kurve muss nur der eingestellte Winkel beibehalten werden, d. h. es ist ein konstantes Drehmoment erforderlich. Dieses wird durch einen annähernd konstanten Öldruck eingestellt. Infolgedessen ist der Ölfluss ungefähr gleich Null. Ein solches Verhalten ist im Stand der Technik zu beobachten, siehe \cite{2016_PracticalTrajectoryDesigns_fodor} \cite{2019_EnergyefficientHighprecisionControl_koivumaki}. 
Zusammenfassend lässt sich sagen, dass das Gesamtmodell für die Modellierung des Fahrzeugmanipulators und für eine allgemeine Analyse der gemeinsamen Steuerungskonzepte geeignet ist.

\subsection{Validierung mit einem menschlichen Bediener}
Anschließend wurde das Gesamtsystem mit einem menschlichen Bediener getestet. Dabei wurde die kooperative Regelung untersucht und mit einem nicht-kooperativen Konzept verglichen. Die kooperative Regelung unterstützt den Bediener in schwierigeren Situationen, in denen der Manipulator nicht mehr den Referenztrajektorien folgen kann. Hingegen folgt die nichtkooperative Regelung nur den vorgegebenen Referenztrajektorien des Fahrzeugs ohne den Menschen zu unterstützen
Die Abb. \ref{fig:thms_validation} zeigt den Vorteil der kooperativen Regelung. Das Fahrzeug verlässt seine Referenz mit der kooperativen Regelung, um den Bediener zu helfen. Damit kann der Bediener mit dem Manipulator die Referenz besser erreichen, siehe $x=30\,$m, $x=75\,$m and $x=100\,$m. Es lässt sich feststellen, dass die Trajektorien mit der kooperativen Regelung besser verfolgt werden können.
Der Simulator ermöglicht die Erprobung in verschiedenen Szenarien. Sogar kritische und gefährliche Versuche sind möglich, ohne solche Maschinen zu beschädigen oder den Menschen zu verletzten. Folglich trägt der Simulator bei, um Kosten zu sparen und die Entwicklungszeit zu verkürzen. Außerdem können auch verschiedene Bedienungskonzepte weitgehend getestet werden, siehe \cite{2020_SharedControlConceptsLarge_varga} oder~\cite{2020_ValidationCooperativeSharedControl_varga}.
\addtolength{\textheight}{-12.5cm} 
\section{Conclusion}
In diesem Beitrag wurde ein Simulator für eine große mobile Arbeitsmaschine vorgestellt, welche sich für die Validierung kooperativer Regelungskonzepte eignet. Die Notwendigkeit des nichtlinearen Fahrzeugmodells ist durch den Einfluss des Manipulators zu erkennen: Das Fahrzeug ist während der Arbeit nach vorne geneigt, weshalb ein nichtlineares Modell die Bewegungen nicht korrekt wiedergeben würde. Die Modelle wurden qualitativ validiert und gezeigt, dass das System realistische Bewegungen solcher Systeme beschreiben kann. Anschließend wurde ein Test präsentiert, in welchem ein menschlicher Bediener das System ansteuert und mit dem untersuchten kooperativen Regelungskonzept interagiert. Es wurde gezeigt, dass mit dem entwickelten Simulator möglich ist, vielfältige Szenarien zu untersuchen und verschiedene Reglereinstellungen zu testen.

\bibliographystyle{IEEEtran}






\end{document}